\pdfoutput=1

\documentclass[11pt]{article}

\usepackage[]{acl}

\usepackage{times}
\usepackage{latexsym}

\usepackage[T1]{fontenc}

\usepackage[utf8]{inputenc}

\usepackage{microtype}
\usepackage{amsmath,enumitem}
\usepackage{graphicx}  

\title{Improving Graph-Based Text Representations with Character and Word Level N-grams}

 \author{Wenzhe Li \and Nikolaos Aletras \\
         Computer Science Department, University of Sheffield, UK\\
         \texttt{\{wli90, n.aletras\}@sheffield.ac.uk}
         }

\begin{document}
\maketitle
\begin{abstract}
Graph-based text representation focuses on how text documents are represented as graphs for exploiting dependency information between tokens and documents within a corpus. Despite the increasing interest in graph representation learning, there is limited research in exploring new ways for graph-based text representation, which is important in downstream natural language processing tasks. In this paper, we first propose a new heterogeneous word-character text graph that combines word and character \textit{n}-gram nodes together with document nodes, allowing us to better learn dependencies among these entities. Additionally, we propose two new graph-based neural models, \texttt{WCTextGCN} and \texttt{WCTextGAT}, for modeling our proposed text graph. Extensive experiments in text classification and automatic text summarization benchmarks demonstrate that our proposed models consistently outperform competitive baselines and state-of-the-art graph-based models.\footnote{Code is available here: \url{https://github.com/GraphForAI/TextGraph}}
\end{abstract}


\section{Introduction}

State-of-the art graph neural network (GNN) architectures~\cite{scarselli2008graph} such as graph convolutional networks (GCNs)~\cite{kipf2016semi} and graph attention networks (GATs)~\cite{velivckovic2017graph} have been successfully applied to various natural language processing (NLP) tasks such as text classification~\cite{yao2019graph,liang2022aspect,ragesh2021hetegcn,yao2021knowledge} and automatic summarization~\cite{wang2020heterogeneous,an2021enhancing}.

\begin{figure*}[!t]
    \centering
    \includegraphics[width=0.90\textwidth]{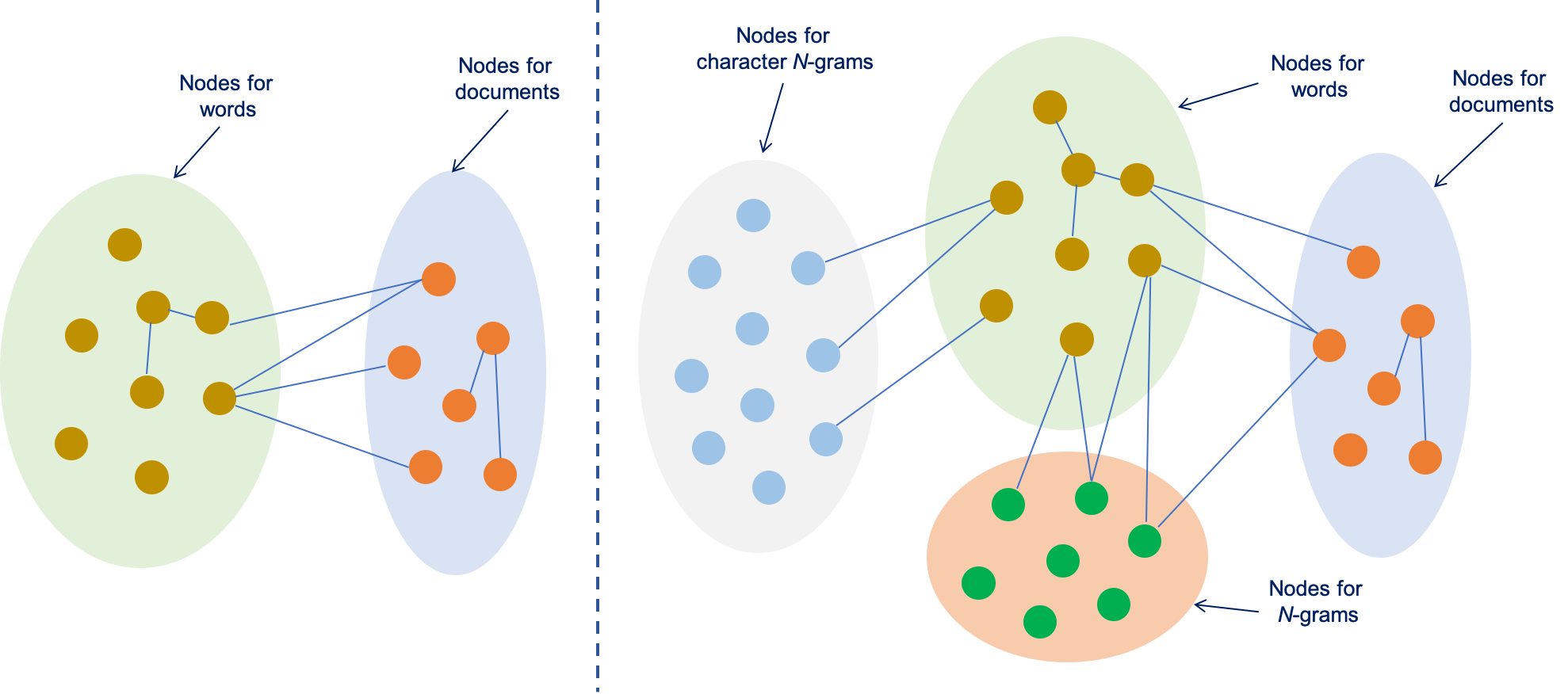}
    \caption{A simple word-document graph (left); and our proposed Word-Character Heterogeneous graph (right). For right figure, the edge types are defined as follows: (1) word-document connection if a document contains a word (i.e., tf-idf); (2) word-word connection based on co-occurrence statistics (i.e., PMI); (3) document-document connection with similarity score (i.e., cosine similarity); (4) word \textit{n}-grams and words connection if a word is part of \textit{n}-grams (0/1); (5) word \textit{n}-grams and document connection if a document contains a n-grams (0/1); and (6) character \textit{n}-grams and words connection if a character \textit{n}-grams is part of a word (0/1). \label{fig:model}}
\end{figure*}

The success of GNNs in NLP tasks highly depends on how effectively  the text is represented as a graph. A simple and widely adopted way to construct a graph from text is to represent documents and words as graph nodes and encode their dependencies as edges (i.e., word-document graph). A given text is converted into a \emph{heterogeneous graph} where nodes representing documents are connected to nodes representing words  if the document contains that particular word~\cite{minaee2021deep,wang2020heterogeneous}. Edges among words are typically weighted using word co-occurrence statistics that quantify the association between two words, as shown in Figure~\ref{fig:model} (left). 

However, word-document graphs have several drawbacks. Simply connecting individual word nodes to document nodes \textit{ignores the ordering of words in the document} which is important in understanding the semantic meaning of text. Moreover, such graphs cannot deal effectively with \textit{word sparsity}. Most of the words in a corpus only appear a few times that results in inaccurate representations of word nodes using GNNs. This limitation is especially true for languages with large vocabularies and many rare words, as noted by~\cite{bojanowski2017enriching}. Current word-document graphs also ignore \textit{explicit document relations} i.e., connections created from pair-wise document similarity, that may play an important role for learning better document representations~\cite{litextsgcn2020}.

\noindent \textbf{Contributions:} In this paper, we propose a new simple yet effective way of constructing graphs from text for GNNs. First, we assume that word ordering plays an important role for semantic understanding which could be captured by higher-order \textit{n}-gram nodes. Second, we introduce character \textit{n}-gram nodes as an effective way for mitigating  sparsity~\cite{bojanowski2017enriching}. Third, we take into account document similarity allowing the model to learn better associations between documents. Figure~\ref{fig:model} (right) shows our proposed Word-Character Heterogeneous text graph compared to a standard word-document graph (left). Finally, we propose two variants of GNNs, \texttt{WCTextGCN} and \texttt{WCTextGAT}, that extend GCN and GAT respectively, for modeling our proposed text graph.


\section{Methodology}
Given a corpus as a list of text documents $\mathcal{C}=\{D_1,...,D_n\}$, our goal is to learn an embedding $\mathbf{h}_i$ for each document $D_i$ using GNNs. This representation can subsequently be used in different downstream tasks such as text classification and summarization.


\subsection{Word-Character Heterogeneous Graph}

The Word-Character Heterogeneous graph $G=(V,E)$ consists of the node set $V=V_d\cup V_w\cup V_{g}\cup V_{c}$, where $V_d=\{d_1,..,d_n\}$ corresponds to a set of documents,  $V_w=\{w_1,...,w_m\}$ denotes a set of unique words, $V_{g}=\{g_1,...,g_l\}$ denotes a set of unique \textit{n}-gram tokens, and finally $V_c=\{c_1,...,c_p\}$ denotes a set of unique character \textit{n}-grams. The edge types among different nodes vary depending on the types of the connected nodes. In addition, we also add edges between two documents if their cosine similarity is larger than a pre-defined threshold.

\subsection{Word and Character \textit{N}-grams Enhanced Text GNNs}

The goal of GNN models is to learn representation for each node. We use $\mathbf{H}^d\in R^{n_d\times k}, \mathbf{H}^w\in R^{n_w\times k}, \mathbf{H}^g\in R^{n_g\times k}, \mathbf{H}^c\in R^{n_c\times k}$ to denote representations of document nodes, word nodes, word \textit{n}-grams nodes and character \textit{n}-grams nodes, where $k$ is the size of the hidden dimension size. $n_d, n_w, n_g, n_w$ represent the number of documents, words, word \textit{n}-grams and character \textit{n}-grams in the graph respectively. We use $e^{dw}_{ij}$ to denote the edge weight between the $i$th document and $j$th word. Similarly, $e^{cw}_{kj}$ denotes the edge weight between the $k$th character \textit{n}-gram and $j$th word. 

The original GCN and GAT models only consider simple graphs where the graph contains a single type of nodes and edges. Since we now are dealing with our Word-Character Heterogeneous graph, we introduce appropriate modifications. 

\paragraph{Word and Character \textit{N}-grams Enhanced Text GCN (\texttt{WCTextGCN})}

In order to support our new graph type for GCNs, we need a modification for the adjacency matrix $\mathbf{A}$. The updating equation for original GCN is:
\[
\mathbf{H}^{(L+1)}=f(\hat{\mathbf{A}}\mathbf{H}^L\mathbf{W}_L)
\]
where $W_L$ is the free parameter to be learned for layer $L$. We assume $\mathbf{H}$ is simply the concatenation of $\mathbf{H}^d,\mathbf{H}^w,\mathbf{H}^g,\mathbf{H}^c$. For \texttt{WCTextGCN}, the adjacency matrix $\mathbf{A}$ is re-defined as:
\begin{equation*}
\mathbf{A} = 
\begin{pmatrix}
\mathbf{A}^{dd}_{sim} & \mathbf{A}^{dw}_{tfidf} & \mathbf{A}^{dg}_{tfidf} & -\\
\mathbf{A}^{wd}_{tfidf} & \mathbf{A}^{ww}_{pmi} & \mathbf{A}^{wg}_{0/1} & \mathbf{A}^{wc}_{0/1}\\
\mathbf{A}^{gd}_{tfidf} & \mathbf{A}^{gw}_{0/1} & - & -\\
- & \mathbf{A}^{cw}_{0/1} & - & -
\end{pmatrix}
\end{equation*}
where $\mathbf{A}^{dd}_{sim}$ denotes the pair-wise similarity between documents~\footnote{We remove edges with similarity score less than a pre-defined threshold to avoid uninformative links.}, sub-matrix $\mathbf{A}^{dw}_{tfidf}$ represents the tf-idf score for all edges linking documents to words, $\mathbf{A}^{wg}_{0/1}$ is the boolean sub-matrix representing whether a word \textit{n}-gram contains a specific word, and so on. The sub-matrix $\mathbf{A}^{dw}_{tfidf}$ is the transpose of sub-matrix $A^{wd}_{tfidf}$.

\paragraph{Word and Character \textit{N}-grams Enhanced Text GAT \texttt{WCTextGAT}}

In GAT, the updates to the node representation is computed by weighting the importance of neighboring nodes. Since our text graph contains four types of nodes, each updating procedure consists of the following four phases (dependency relation among nodes can be seen in Figure~\ref{fig:model}):
\begin{align}
    \hat{\mathbf{H}}^d &= \text{GAT}(\mathbf{H}^d, \mathbf{H}^w, \mathbf{H}^g) \nonumber\\ 
    \hat{\mathbf{H}}^w &= \text{GAT}(\mathbf{H}^d, \mathbf{H}^w, \mathbf{H}^g, \mathbf{H}^c) \nonumber\\
    \hat{\mathbf{H}}^g &= \text{GAT}(\mathbf{H}^d, \mathbf{H}^w) \nonumber\\ 
    \hat{\mathbf{H}}^c &= \text{GAT}(\mathbf{H}^w) \nonumber 
\end{align}
For example, to update word representation $\hat{\mathbf{H}}^w$, we need to aggregate information from document nodes, word nodes, word \textit{n}-gram nodes and character \textit{n}-gram nodes, respectively. Assume that we update the embedding for word node $i$ by considering neighboring document nodes only (similar procedure applies to other three types of nodes). The computation is as follows:

\begin{align}
    z_{ij}&=\text{Leaky}(a^{T}[\mathbf{W}_v\mathbf{h}^w_i; \mathbf{W}_d\mathbf{h}^d_j; \mathbf{W}_e{e}_{ij}^{wd}])\nonumber\\
    \alpha_{ij}&=\frac{\exp(z_{ij})}{\sum_{l\in \mathcal{N}_i}^{}\exp(z_{il})}\nonumber\\
    \hat{\mathbf{h}^{1}_i}&=\sigma(\sum_{j\in \mathcal{N}_i}^{}\alpha_{ij}\mathbf{W}_d\mathbf{h}^d_j) \nonumber
\end{align}
where $\mathbf{W}_v, \mathbf{W}_d, \mathbf{W}_e$ are the trainable weights of the model, that are applied to different types of nodes. $\alpha_{ij}$ is the attention weight between word $i$ and document $j$. $\mathcal{N}_i$ denotes the set of neighboring documents for word $i$, and $\sigma(.)$ is the activation function. Multi-head attention~\cite{vaswani2017attention} is also introduced to capture different aspects of semantic representations for text:
\begin{equation*}
    \hat{\mathbf{h}^{1}_i}=\parallel_{k=1}^{K}\sigma(\sum_{j\in \mathcal{N}_i}^{}\alpha^k_{ij}\mathbf{W}^k_d\mathbf{h}_j)
\end{equation*}

Similarly, we can also compute $\hat{\mathbf{h}^{2}_i},\hat{\mathbf{h}^{3}_i},\hat{\mathbf{h}^{4}_i}$ by considering other types of neighboring nodes. Finally, these representations are concatenated, followed by linear transformation.




\begin{table*}[]
\centering
\begin{tabular}{|c|ccccc|l}
\cline{1-6}
 Dataset   & \textbf{20NG}     & \textbf{ R8}      & \textbf{ R52}  & \textbf{ Ohsumed}  & \textbf{MR}   \\\cline{1-6}
 \textbf{TF-IDF+LR} & 83.19$\pm$0.00          & 93.74$\pm$0.00         & 86.95$\pm$0.00         & 54.66$\pm$0.00   &    74.59$\pm$0.00\\
 \textbf{fastText} &  79.38$\pm$0.30  &96.13$\pm$0.21 & 92.81$\pm$0.09 & 57.70$\pm$0.49& 75.14$\pm$0.20\\
 \textbf{CNN-rand} & 76.83$\pm$0.61 & 94.02$\pm$0.57 & 85.37$\pm$0.47 & 43.87$\pm$1.00 & 74.98$\pm$0.70 \\
\textbf{CNN-non-static} & 82.15$\pm$0.52 & 95.71$\pm$0.52 & 87.59$\pm$0.48  & 58.44$\pm$1.06&77.75$
\pm$0.72  \\
\textbf{LSTM-rand} & 65.71$\pm$1.52 & 93.68$\pm$0.82 & 85.54$\pm$1.13 & 41.13$\pm$1.17   &75.06$\pm$0.44\\
\textbf{LSTM-pretrain} & 75.43$\pm$1.72          &    96.09$\pm$0.19       & 90.48$\pm$ 0.86          &51.10$\pm$1.50 &77.33$\pm$0.89       \\
\textbf{PTE} &  76.74$\pm$0.29        & 96.69$\pm$ 0.13          & 90.71$\pm$ 0.14       &53.58$\pm$ 0.29  &    70.23$\pm$0.36     & \\
\textbf{BERT} &  83.41$\pm$0.20        & 96.98$\pm$ 0.08          & 92.87$\pm$ 0.01       &67.22$\pm$ 0.27  &    77.02$\pm$0.23     & \\\cline{1-6}

\textbf{TextGCN} &  86.34$\pm$0.09         &97.07$\pm$ 0.10          &93.56$\pm$0.18       & 68.36$\pm$0.56  &  76.74$\pm$0.20        \\
\textbf{WCTextGCN} (Ours) &   \textbf{87.21$\pm$0.54}       &  \textbf{97.49$\pm$0.20}        & 93.88$\pm$0.34      & \textbf{68.52}$\pm$0.20 &  \textbf{77.85$\pm$ 0.34}     \\\cline{1-6}
\textbf{TextGAT} &  85.78$\pm$ 0.10         & 96.88$\pm$0.24         & 93.61$\pm$0.12      &67.46$\pm$0.32  & 76.45$\pm$0.38   \\
\textbf{WCTextGAT} (Ours) &    87.02$\pm$ 0.32  & 97.12$\pm$0.42       & \textbf{94.02$\pm$0.45}        & 68.14$\pm$0.18 & 77.98$\pm$0.10         &\\\cline{1-6}
\end{tabular}
\caption{Predictive test accuracy on five text classification benchmark datasets. We run models 10 times and report mean$\pm$standard deviation. 
}
\label{tab:result2}
\end{table*}

\section{Experiments and Results}
We conduct experiments on two NLP tasks, i.e., text classification and extractive summarization. The latter one can be also viewed as a classification problem for each sentence level (i.e., to be included in the summary or not). 

\subsection{Text Classification}

\paragraph{Data} 
We select five widely used benchmark datasets including 20-Newsgroups, Ohsumed, R52, R8 and MR. The statistics and the descriptions for these datasets can be found in~\cite{yao2019graph}.

\paragraph{Baselines}
We compare our models to multiple existing state-of-the-art text classification methods including \textbf{TF-IDF+LR}, \textbf{fastText}~\cite{joulin2016bag}, \textbf{CNN}~\cite{le2014distributed}, \textbf{LSTM}~\cite{liu2016recurrent}, \textbf{PTE}~\cite{tang2015pte}, \textbf{BERT}~\cite{devlin2018bert}, \textbf{TextGCN}~\cite{yao2019graph} and \textbf{TextGAT}.

\begin{table*}[!t]
\centering
\small
\begin{tabular}{|c|cccc|l}
\cline{1-5}
R8    & 3     & 4     & 5  &6    \\\cline{1-5}
3  &97.1&\textbf{97.5}&\textbf{97.5}&97.4 \\
4  &&96.9&97.1&97.5 \\
5  &&&97.1&97.4 \\
6  &&&&97.4 \\\cline{1-5}
\end{tabular}
\quad
\begin{tabular}{|c|cccc|l}
\cline{1-5}
  R52  & 3     & 4     & 5  &6    \\\cline{1-5}
3  &93.5&\textbf{93.8}&93.8& 93.7 \\
4  &&93.4&93.6& \textbf{93.8}\\
5  &&&93.6&93.7 \\
6  &&&&\textbf{93.8} \\\cline{1-5}
\end{tabular}
\quad
\begin{tabular}{|c|cccc|l}
\cline{1-5}
  MR  & 3     & 4     & 5  &6    \\\cline{1-5}
3  &76.8&78.2&\textbf{78.3}&\textbf{78.3} \\
4  &&77.2&78.2& \textbf{78.3} \\
5  &&&78.1& 78.1 \\
6  &&&& 77.9\\\cline{1-5}
\end{tabular}
\caption{The effect on performance by using character \textit{n}-grams of $n$ in \{3,..,6\}.}
\label{tab:result3}
\end{table*}

\paragraph{Experimental Settings}
We randomly select 10\% of the training set for the validation. For the \texttt{WCTextGCN} model, we set the hidden size to $200$. For the TextGAT and \texttt{WCTextGAT} models, we use $8$ attention heads with each containing $16$ hidden units, and set edge feature dimension to $32$. The learning rate is $0.002$ and dropout rate $0.5$. We train all models for $200$ epochs using Adam optimizer~\cite{kingma2014adam} and early stopping with patience $20$. For all the GNNs models, we use two hidden layers and $1$-of-$K$ encoding for initialization.

\paragraph{Results}
Table~\ref{tab:result2} shows the text classification results. We observe that the incorporation of word \textit{n}-grams, character \textit{n}-grams and document similarity are helpful and consistently improve predictive performance over other models. i.e., the \texttt{WCTextGCN} model improves accuracy on 20NG over 0.8\% compared to the TextGCN model. The improvements in MR and R8 datasets are more substantial than others, 0.5\% and 1.1\%, respectively. This is because character \textit{n}-grams help more when text is short, which is consistent with our hypothesis that character \textit{n}-grams are helpful for mitigating sparsity problems.

\paragraph{Varying the size of \textit{n}-grams}
For character \textit{n}-grams, we set \textit{n}-grams ranging from $3$ to $6$ characters, and record the performance in different combinations of \textit{n}-grams, i.e., $3$-grams to $4$-grams, $3$-grams to $5$-grams and so on. The results are shown in Table~\ref{tab:result3} with best scores in bold. We observe that the best results are often obtained when we vary the range of $n$ from $3$ to $4$. Further increase of $n$ provides limited effects in model performance. In terms of word $n$-grams, we observe similar results.

\subsection{Extractive Text Summarization}
Extractive single-document summarization is formulated as a binary classification for each sentence with the aim to predict whether a sentence should be included in the summary or not. We follow the same setting as the HeterogeneousSumGraph (HSG) proposed by~\citet{wang2020heterogeneous} except that we use our new Word-Character Heterogeneous graph representation denoted as \texttt{HSG-Ours}. 

\paragraph{Data}
We select two widely used benchmark newes articles datasets, \textit{CNN/DailyMail}~\citep{hermann2015teaching} and \textit{NYT50}~\cite{durrett2016learning}. The first contains 
287,227/13,368/11,490 examples for training, validation and test. The second contains 110,540 articles with their summaries and is split into 100,834 and 9,706 for training and test. Following~\citet{durrett2016learning}, we use the last 4,000
documents from the training set for validation and 3,452 test examples.

\paragraph{Baselines and Experimental Settings}
We evaluate our models on single document summarization by comparing to three different baselines~\cite{wang2020heterogeneous}, Ext-BILSTM, Ext-Transformer and HSG. For all experiments, we simply follow the same settings as in~\citet{wang2020heterogeneous} and evaluate performance using ROUGE~\cite{lin2003automatic}.

\paragraph{Results}
Tables~\ref{tab:results4} and \ref{tab:results5} show the ROUGE scores on the two datasets. \texttt{HGS-Ours} with our new text graph performs consistently better than competing ones. In particular, for NYT50 data, the R-1 and R-2 metrics improve more than $0.5$ compared to the HSG model. We observe similar performance differences for R-L on CNN/DailyMail data. This highlights the efficacy of our new text graph in learning better word and sentence representations, especially for the words that appear only few times but play an important role in summarization.



\begin{table}[!t]
\centering
\small
\begin{tabular}{|c|ccc|l}
\cline{1-4}
 Model   & \textbf{R-1}     & \textbf{ R-2}      & \textbf{ R-L}    \\\cline{1-4}
 \textbf{Ext-BiLSTM} & 46.32         & 25.84         & 42.16    \\
  \textbf{Ext-Transformer} & 45.07        & 24.72         & 40.85     \\
 \textbf{HSG} & 46.89         & \textbf{26.26}         & 42.58     \\
 \textbf{HSG-Ours} &    \textbf{46.96}      & 26.20         & \textbf{43.43}     \\\cline{1-4}
\end{tabular}
\caption{Performance (ROUGE) of different models on CNN/DailyMail.}
\label{tab:results4}
\end{table}

\begin{table}[]
\small
\centering
\begin{tabular}{|c|ccc|l}
\cline{1-4}
 Model   & \textbf{R-1}     & \textbf{ R-2}      & \textbf{ R-L}    \\\cline{1-4}
 \textbf{Ext-BiLSTM} & 41.59        & 19.03         & 38.04    \\
  \textbf{Ext-Transformer} & 41.33        & 18.83         & 37.65     \\
 \textbf{HSG} & 42.31         & 19.51         & 38.74     \\
 \textbf{HSG-Ours} & \textbf{42.85}         & \textbf{20.03}         & \textbf{38.90}     \\\cline{1-4}
\end{tabular}
\caption{Performance (ROUGE) of different models on NYT50.}
\label{tab:results5}
\end{table}


\section{Conclusion}
In this paper, we proposed a new text graph representation by incorporating word and character level information. GNN models trained using our text graph provide superior performance in text classification and single-document summarization compared to previous work. In the future, we plan to extend our proposed method to other tasks such as opinion extraction~\cite{mensah-etal-2021-empirical}, misinformation detection~\cite{chandra2020graph,10.7717/peerj-cs.325,10.1145/3501247.3531559}, voting intention forecasting~\cite{tsakalidis2018nowcasting} and socioeconomic attribute analysis~\cite{aletras2018predicting}. We finally plan to extend our GNN models by weighting the contribution of neighboring nodes~\cite{zhang2022node}.
\bibliography{anthology,custom}
\bibliographystyle{acl_natbib}

\end{document}